\documentclass[letterpaper, 10 pt, conference]{ieeeconf}  

\IEEEoverridecommandlockouts                              

\usepackage{graphics} 
\usepackage{epsfig} 
\usepackage{mathptmx} 
\usepackage{times} 
\usepackage{amsmath} 
\usepackage{amssymb}  
\usepackage{pdfpages}
\usepackage{amsfonts}

\let\labelindent\relax
\usepackage{enumitem}

\usepackage{url}
\usepackage{pifont}
\newcommand{\cmark}{\ding{51}}%
\newcommand{\xmark}{\ding{55}}%

\usepackage{xcolor}
\usepackage{multirow}

\usepackage{etoolbox}
\makeatletter
\patchcmd{\@verbatim}
  {\verbatim@font}
  {\verbatim@font\small}
  {}{}
\makeatother

\title{\LARGE \bf
Alternative Modes of Interaction in Proximal\\Human-in-the-Loop Operation of Robots  
}

\author{Tathagata Chakraborti$^{1}$ \and Sarath Sreedharan$^{1}$ \and Anagha Kulkarni$^{1}$ \and Subbarao Kambhampati$^{1}$
\thanks{$^{1}$Tathagata Chakraborti, Sarath Sreedharan, Anagha Kulkarni and Subbarao Kambhampati are with the Department of Computer Science, Arizona State University, Tempe, AZ 85281, USA 
        {\tt\small \{ tchakra2, ssreedh3, akulka16, rao \} @ asu.edu }}%
\thanks{The work is part of {\em Project Cloudy with a Chance of Synergy}, from team \AE Robotics to appear in the US Finals of the Microsoft Imagine Cup 2017 in April. {\tt\small http://www.ae-robots.com/}
}
}

\begin{document}

\maketitle
\thispagestyle{empty}
\pagestyle{empty}


\begin{abstract}

Ambiguity and noise in natural language instructions create a significant barrier towards adopting autonomous systems into safety critical workflows involving humans and machines. In this paper, we propose to build on recent advances in electrophysiological monitoring methods and augmented reality technologies, to develop alternative modes of communication between humans and robots involved in large-scale proximal collaborative tasks. 
We will first introduce augmented reality techniques for projecting a robot's intentions to its human teammate, who can interact with these cues to engage in real-time collaborative plan execution with the robot.
We will then look at how electroencephalographic (EEG) feedback can be used to monitor human response to both discrete events, as well as longer term affective states while execution of a plan. These signals can be used by a learning agent, a.k.a an affective robot, to modify its policy.
We will present an end-to-end system capable of demonstrating these modalities of interaction.
We hope that the proposed system will inspire research in augmenting human-robot interactions with alternative forms of communications in the interests of safety, productivity, and fluency of teaming, particularly in engineered settings such as the factory floor or the assembly line in the manufacturing industry where the use of such wearables can be enforced.

\end{abstract}

\section{Introduction}

\begin{figure}[!ht]
\centering
\includegraphics[width=0.9\columnwidth]{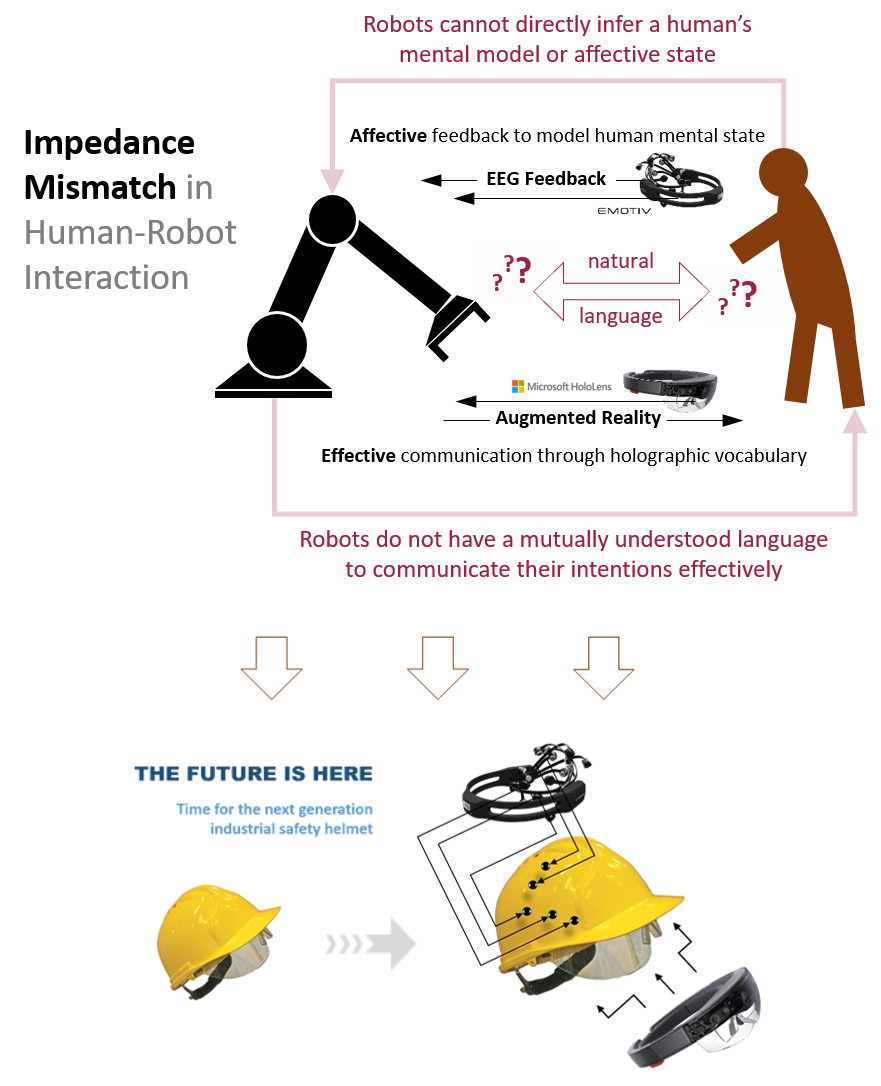}
\caption{
Alternative forms of communication to combat impedance mismatch in human robot interactions in settings where  wearables can be integrated for closed loop feedback from EEG signals and augmented reality. 
}
\label{motivation}
\end{figure}

\noindent 



The last decade has seen a massive increase in robots deployed on the factory floor \cite{robo}. This has led to fears of massive loss of jobs for humans in the manufacturing industry, as well concerns of safety for the jobs that do remain. The latter is not an emerging concern, though.  Automation of the manufacturing industry has gone hand in hand with incidents of misaligned intentions between the robots and their humans co-workers, leading to at least four instances of fatality \cite{brownhri}. 
This dates back to as early as 1979 when a robot arm crushed a worker to death while gathering supplies in the Michigan Ford Motor Factory, to as recent as 2015 in a very similar and much publicized accident in the Volkswagen factory in Baunatal, Germany.
With 1.3 million new robots predicted to enter the workspace by next year \cite{ifr}, such concerns are only expected to escalate.

A closer look at the dynamics of employment in the manufacturing industry also reveals that the introduction of automation has in fact increased productivity \cite{hbr} as well as, surprisingly, contributed to a steady increase in the number of jobs for human workers \cite{bloomberg} in Germany (which so far dominates in terms of deployed robots in the industry). 
We posit then either a semi-autonomous workspace in future with increased hazards due to misaligned interests of robots in the shared environment, or a future where the interests of the human workers will be compromised in favor of automation.
In light of this, it is essential that the next-generation factory floor is able to cope with the needs of these new technologies. 

At the core of this problem is the impedance mismatch between humans and robots in how they communicate, as illustrated in Figure \ref{motivation}. 
Despite the progress made in natural language processing, natural language understanding is still a largely unsolved problem, and as such robots find it difficult to {\bf (1)} express their own goals and intentions effectively; as well as {\bf (2)} understand human expressions and emotions. Thus there exists a significant communication barrier to be overcome from either side, and robots are essentially still ``autistic" \cite{kaminka2013curing} in many aspects. 
While this may not be a serious concern for deploying completely autonomous agents in isolated environments such as for space or underwater exploration, the priorities change considerably when humans and robots are involved in collaborative tasks, especially for concerns of safety, if not to just improve the effectiveness of collaboration.
This is also emphasized in the \emph{Roadmap for U.S. Robotics} report, which outlines that {\em ``humans must be able to read and recognize robot activities in order to interpret the robot's understanding''} \cite{christensen2009roadmap}. 
Recent work on this has focused on generation of legible robot motion planning \cite{Dragan2015a} and explicable task planning \cite{exp-yu}, as well as verbalization of robot intentions using natural language~\cite{Tellex-RSS-14,16roman-verbalization}.

\subsection*{The Manufacturing Environment.} 

Our primary focus here is on structured settings like the manufacturing environment where wearables can be a viable solution for improving the workspace. Indeed, a reboot of the safety helmet and goggles as illustrated in Figure \ref{motivation} only requires retro-fitting existing wearables with sensors that can enable these new technologies. 
Imagine, then, a human and robot engaged in an assembly task, where they are constructing a structure collaboratively. Further suppose that the human now needs a tool from the shared workspace. At this time, neither agent is sure what tools and objects the other is going to access in the immediate future - this calls for seamless transfer of relevant information without loss of workflow. Existing (general purpose) solutions will suggest intention recognition \cite{hayes} or natural language \cite{Tellex-RSS-14} communication as a means to respond to this situation. 
With regards to naturalistic modes of interaction among agents, while natural language and intent or gesture recognition techniques remain the ideal choice in most cases, and perhaps the only choice in some (such as robots that would interact with people in their daily lives), 
we note that these are inherently noisy and ambiguous, and not necessary in controlled environments such as on the factory floor or by the assembly line where the workspace can be engineered to enforce protocols in the interests of safety and productivity, in the form of safety helmets integrated with wearable technology \cite{ruffaldi2016third}.

Thus, in our system, the robot instead projects its intentions as {\em holograms} thus making it directly accessible to the human in the loop, e.g. by projecting a pickup symbol on a tool it might use in future. Further, unlike in traditional mixed reality projection systems, the human can directly interact with these holograms to make his own intentions known to the robot, e.g. by gazing at and selecting the desired tool thus forcing the robot to replan. To this end, we develop, with the power of the HoloLens\footnotemark, an alternative communication paradigm that is based on the projection of explicit visual cues pertaining to the plan under execution via holograms such that they can be intuitively understood and directly read by the human partner. 
The ``real" shared human-robot workspace is now thus augmented with the virtual space where the physical environment is used as a medium to convey information about the intended actions of the robot, the safety of the work space, or task-related instructions. We call this the {\em Augmented Workspace}. Recent development of augmented reality techniques \cite{verge} has opened up endless possibilities in such modes of communication.

This, by itself, however, provides little indication of the mental state of the human, i.e. how he is actually responding to the interactions - something that human teammates naturally keep track of during a collaborative exercise. In our system, we propose to use real-time EEG feedback using the Emotiv EPOC+ headset\footnotemark this purpose. This has several advantages - specific signals in the brain are understood to have known semantics (more on this later), and are detected immediately and with high accuracy, thus short circuiting the need for the relatively highly inaccurate and slower signal processing stage in rivaling techniques such as emotion and gesture recognition. Going back to our previous use case, if the robot now makes an attempt to pick up the same tool again, the error can fire an event related EEG response - which may readily be used as in a closed loop feedback to control or stop the robot. Further, if the robot is making the same mistake again and again, causing the human to be stressed and/or irritated, it can listen to the human's affective states to learn better, and more human-aware, policies over time. We demonstrate these capabilities as part of the {\em Consciousness Cloud} which provides the robots real-time shared access to the mental state of all the humans in the workspace. 
The agents are thus able to query the cloud about particulars (e.g. stress levels) of the current mental state, or receive specific alerts related to the human's response to events (e.g. oddball incidents like safety hazards and corresponding ERP spikes) in the environment. 

Finally, instead of the single human and robot collaborating over an assembly task, imagine now an entire workspace shared by many such agents, as is the case of most manufacturing environments. Traditional notions of communication become intractable in such settings. With this in mind, we make the entire system cloud based - all the agents log their respective states on to a central serve, and can also access the state of their co-workers from it. As opposed to peer-to-peer information sharing, this approach provides a distinct advantage towards making the system scalable to multiple agents, both humans and robots, sharing and collaborating in the same workspace, as envisioned in Figure \ref{overview}.
\footnotetext[1]{https://www.microsoft.com/microsoft-hololens/en-us}
\footnotetext[2]{https://www.emotiv.com/epoc/}

\subsection*{Contributions.} 

Thus, in this paper, we propose approaches to tear down the communication barrier between human and robot team members {\bf (1)} by means of holograms/projections as part of a shared alternative vocabulary for communication in the {\bf Augmented Workspace}, and {\bf (2)} by using direct feedback from physiological signals to model the human mental state in the shared {\bf Consciousness Cloud}.
The former allows for real-time interactive plan monitoring and execution of the robot with a human-in-the-loop, while the latter, in addition to passive plan monitoring, also allows a planning agent to learn preferences of its human co-worker and update its policies accordingly. 
We will demonstrate how this can be achieved on an end-to-end cloud-based platform built specifically to scale up to the demands of the next-generation semi-autonomous workspace envisioned in Figure \ref{overview}. 

\section{Related Work}

\subsection{Intention Projection and Mixed Reality} 

The concept of intention projection for autonomous systems have been explored before. An early attempt was made by \cite{sato2000human} in their prototype Interactive Hand Pointer (IHP) to control a robot in the human's workspace. Similar systems have since been developed to visualize trajectories of mobile wheelchairs and robots \cite{watanabe2015communicating,chadalavada2015s}, which suggest that humans prefer to interact with a robot when it presents its intentions directly as visual cues. The last few years have seen active research \cite{omidshafiei2015mar,omidshafiei2016measurable,shen2013multi,ishii2009designing,mistry2010blinkbot,leutert2013spatial} 
in this area, but most of these systems were passive, non-interactive and quite limited in their scope, and did not consider the state of the objects or the context of the plan pertaining to the action while projecting information. 
As such, the scope of intention projection has remained largely limited. Instead, in this paper, we demonstrate a system that is able to provide much richer information to the human-in-the-loop during collaborative plan execution, in terms of the current state information, action being performed as well as future parts of the plan under execution. We also demonstrate how recent advances in the field of augmented reality make this form of online interactive plan execution particularly compelling. In Table \ref{table-vs} we provide the relative merits of augmented reality with the state-of-the-art in mixed reality projections.

\subsection{EEG Feedback and Robotics}

Electroencephalography (EEG) is an electrophysiological monitoring method to measure voltage fluctuations resulting from ionic currents within the brain. The use of EEG signals in the design of BCI has been of considerable interest in recent times. The aim of our project is to integrate EEG-based feedback in human-robot interaction or HRI.
Of particular interest to us are Event Related Potentials or ERPs which are measured due the response to specific sensory, cognitive, or motor events, and may be especially useful in gauging the human reaction to specific actions during the execution of a robot's plan \cite{hoffmann2008efficient,acqualagna2010novel,guger2012many,ramli2015using}.
Recently, researchers have tried to improve performance in robotics tasks by applying error-related potentials or ErrPs \cite{rao2013brain,ferrez2008error} to a reinforcement learning process \cite{iturrate2010single,iturrate2012latency}. These are error signals produced due to undesired or unexpected effects after performing an action. The existence of ErrPs and the possibility of classifying them in online settings has been studied in driving tasks \cite{zhang2015eeg}, as well as to change the robot’s immediate behavior \cite{sad}. 
However, almost all of the focus has remained on the control of robots rather than as a means of learning behavior \cite{6461528}, and very little has been made of the effect of such signals on the task level interactions between agents. This remains the primary focus of our system.

\begin{figure}[tbp!]
\centering
\includegraphics[width=\columnwidth]{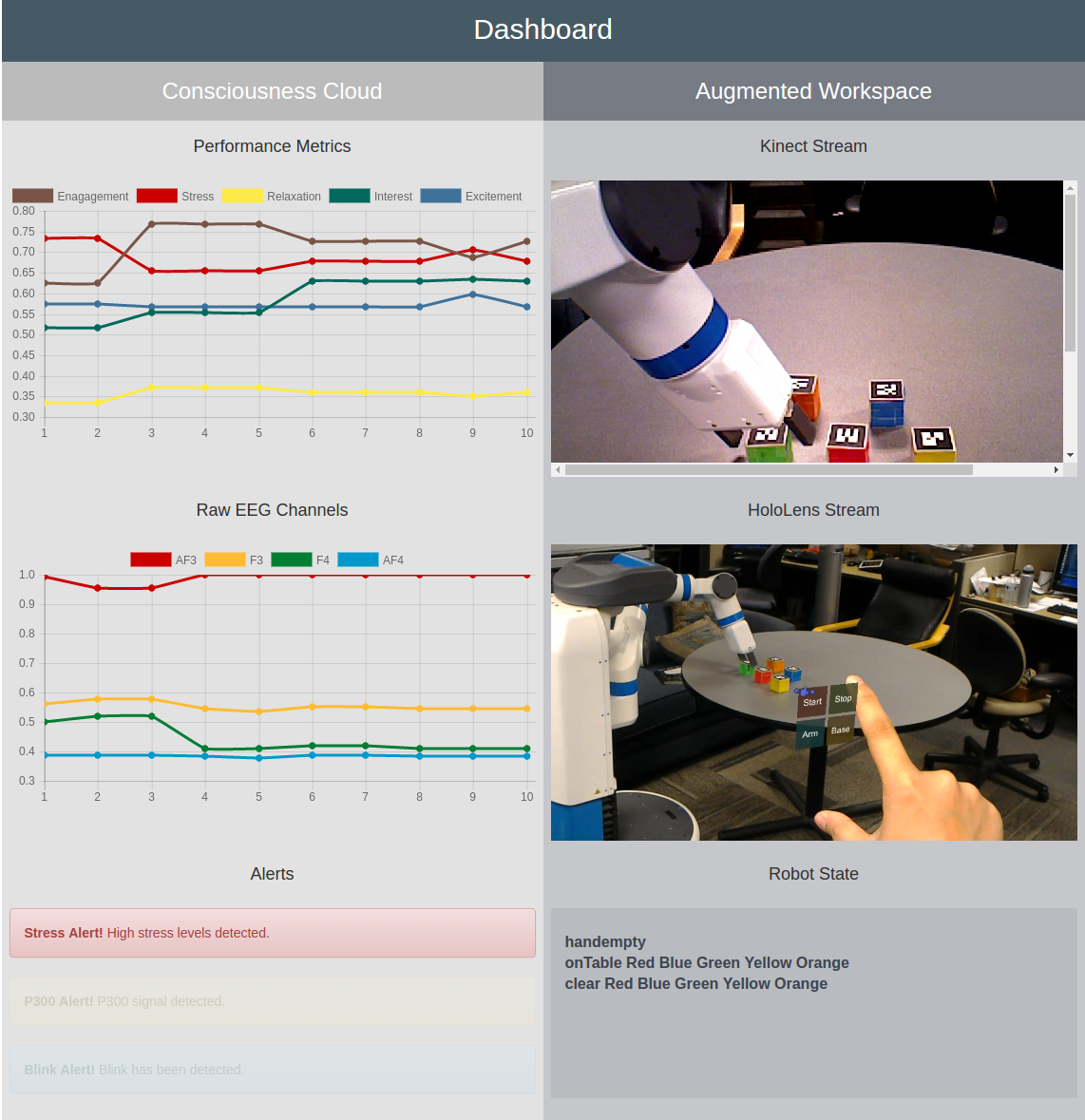}
\caption{
The Dashboard - displaying elements of the Consciousness Cloud and the Augmented Workspace - monitors the state of the shared workspace. 
}
\label{dashboard}
\end{figure}

\begin{figure*}
\centering
\includegraphics[width=\textwidth]{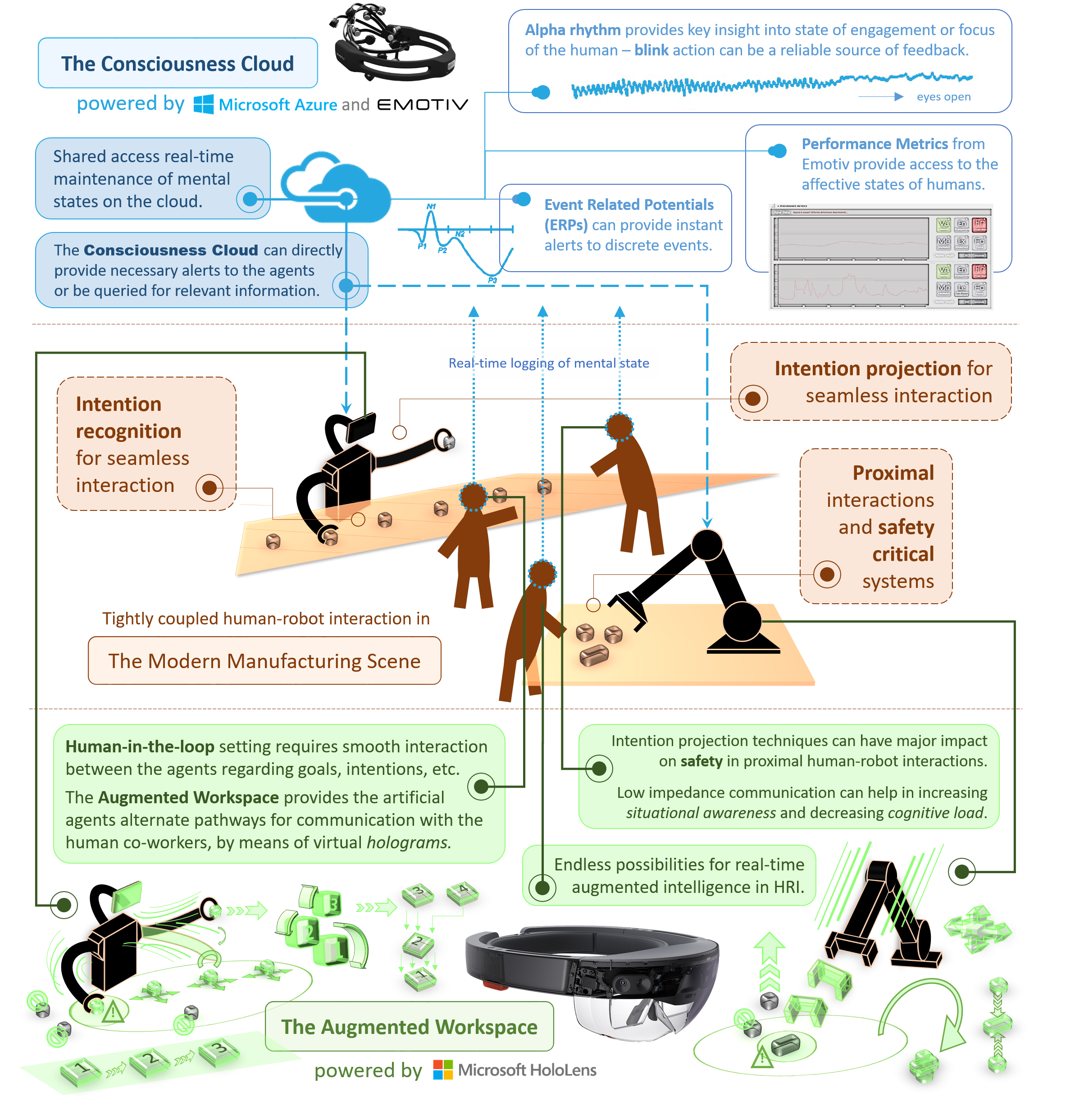}
\caption{
A conceptual impression of the next generation workshop floor involving multiple humans and robots sharing the workspace and collaborating either singly or in groups across different tasks. The humans are wearing safety helmets integrated with electrodes to capture EEG feedback, as well as HoloLens style safety goggles that provide access to augmented reality based communication through a shared-access cloud platform. This means that all the robots now have access to a real-time mental model of all their human co-workers on the cloud which they can use to inform or modulate their own behavior. The robots can also project their goals and intentions, as well as their private regions of interest, into their immediate environment, thereby the improving situational awareness of their human teammates. These two components - called the {\em Consciousness Cloud} and the {\em Augmented Workspace} -  forms a sophisticated plan execution and plan monitoring system that can adapt while taking real-time feedback from humans-in-the-loop.
}
\label{overview}
\end{figure*}

\section{System Overview}

There are two major components of the system (refer to Figure \ref{overview}) - (1) the {\bf Augmented Workspace} which allows the robots to communicate with their human co-workers in the virtual space; and  (2) the {\bf Consciousness Cloud} which provides the robots real-time shared access to the mental state of all the humans in the workspace. 
This is visible in the centralized {\bf Dashboard} that provides a real-time snapshot of the entire workspace, as seen in Figure \ref{dashboard}.
The Augmented Workspace Panel shows real-time stream from the robot's point of view, the augmented reality stream from the human's point of view and information about the current state of plan execution.
The Consciousness Cloud Panel displays the real-time affective states (engagement, stress, relaxation, excitement and interest), raw EEG signals from the four channels (AF3, F3, AF4 and F4) used to detect response to discrete events, as well as alerts signifying abnormal conditions (p300, control blink, high stress, etc.).
The Dashboard allows the humans to communicate or visualize the collaborative planning process between themselves. It can be especially useful in factory settings to the floor manager who can use it to effectively monitor the shared workspace. We will now go into the implementation and capabilities of these two components in more detail.

\begin{figure}[tbp!]
\centering
\includegraphics[width=0.75\columnwidth]{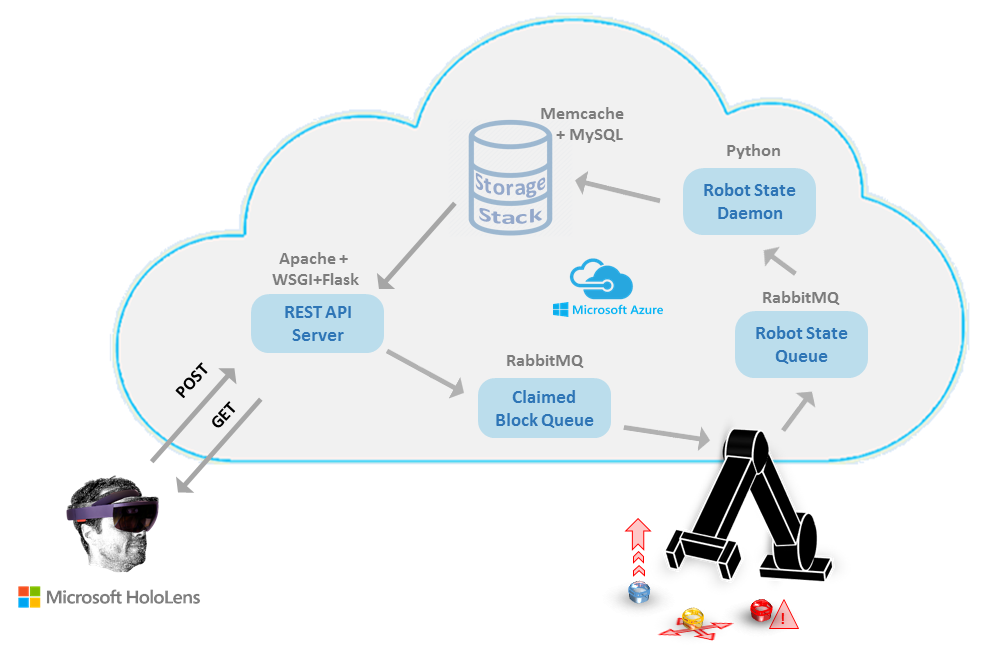}
\caption{Architecture diagram of the Augmented Workspace.}
\label{aw-arch}
\end{figure}

\begin{table*}[!ht]
\begin{center}
    \begin{tabular}{| c || c | c || p{13cm} |}
    \hline
    {\bf Property} & {\bf AR} & {\bf MR} & {\bf Comments} \\ \hline \hline
    \multirow{3}{*}{Interaction} & \multirow{3}{*}{\textcolor{green}{\cmark}} & \multirow{3}{*}{\textcolor{red}{\xmark}} & 
{\small One of the key features of AR is that it provides the humans with the ability to interact directly, and effectively, with the holograms. This becomes particularly difficult in MR, especially due to difficulties in accurate gaze and gesture estimation.}
    \\ \hline
    \multirow{3}{*}{Occlusion} & \multirow{3}{*}{\textcolor{blue}{{\bf ?}}} & \multirow{3}{*}{\textcolor{red}{\xmark}} &
{\small Unlike MR, AR is not particularly disadvantaged by occlusions due to objects or agents in the workspace. However, it does reduce the field of view significantly (though this is expected to improve with future iterations of the HoloLens).}
    \\ \hline
    \multirow{4}{*}{Ergonomics} & \multirow{4}{*}{\textcolor{red}{\xmark}} & \multirow{4}{*}{\textcolor{green}{\cmark}} & 
{\small At present the size, weight and the occlusion of the peripheral view due to the HoloLens makes it somewhat unsuitable for longer operations, while the MR approach does not require any wearables and leaves the human mostly uninhibited. However, this is again expected to improve in later iterations of the HoloLens, as well as if they are custom made and optimized for a setting such as this.}
    \\ \hline
    \multirow{3}{*}{Scalability} & \multirow{3}{*}{\textcolor{green}{\cmark}} & \multirow{3}{*}{\textcolor{blue}{{\bf ?}}} &
{\small MR will find it difficult to scale up to beyond peer-to-peer interactions or a confined space, given the requirement of viable projectors for every interaction. This is hardly an issue for the HoloLens which provides unrestricted mobility and portability of solutions.}
    \\ \hline
    \multirow{3}{*}{Scope} & \multirow{3}{*}{\textcolor{green}{\cmark}} & \multirow{3}{*}{\textcolor{red}{\xmark}} &
{\small MR is limited by a 2D canvas (environment), whereas AR can not only provide 3D projections that can be interacted with but also can express information that 2D projections cannot - e.g. a 3D volume of safety around the robot rather than just the projected area on the floor.}
    \\ \hline \hline
    \end{tabular}
\end{center}
\caption{
Relative merits of Augmented Reality (AR) and Existing Mixed Reality (MR) approaches for intention projection.}
\label{table-vs}
\end{table*}

\section{The Augmented Workspace}

In the augmented workspace (refer to Figure \ref{aw-arch}). the HoloLens communicates with the user endpoints through the \texttt{REST API server}. The API server is implemented in python using the \texttt{Flask} web server framework.  All external traffic to the server is handled by an \texttt{Apache2} server that communicates with the python application through a \texttt{WSGI} middle layer. The \texttt{Apache2} server ensures that the server can easily support a large number of concurrent requests. 

The \texttt{REST} service exposes both \texttt{GET} and \texttt{POST} endpoints. The \texttt{GET} links provides the HoloLens application with a way of accessing information from the robot, while the \texttt{POST} link provides  the  HoloLens  application  control  over  the  robot’s  operation.  Currently, we are using the API to expose information like the robotic planning state, robot joint values and transforms to special markers in the environment. Most  API GET calls will first  try  to  fetch  the requested information from the memcached layer, and would only try a direct query to the \texttt{MySQL} database if the  cache  entry  is  older  than  a  specified limit. Each query to the database also causes the corresponding cache entry to be updated. The \texttt{MySQL} server itself is updated by a daemon that runs on \texttt{Azure} and keeps consuming messages sent from the robot through various queues implemented using the rabbitMQ service.

\begin{figure*}
\centering
\includegraphics[width=0.9\textwidth]{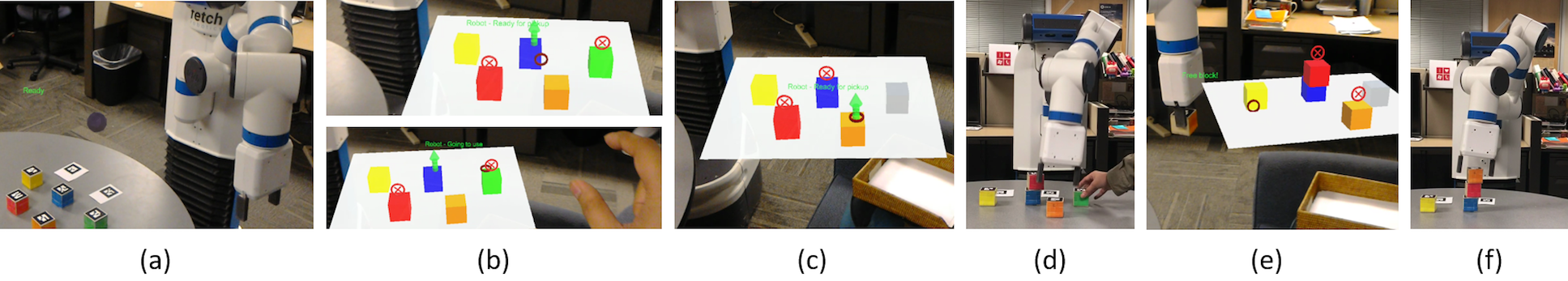}
\caption{
Interactive execution of a plan in the augmented workspace - {\bf (a)} First person view of the real workspace showing initial state. The robot wants to build a tower of height three with blocks blue, red and green. {\bf (b)} Block are annotated with intuitive holograms, e.g. an upward arrow on the block the robot is going to pick up immediately and a red cross mark on the ones it is planning to use later. The human can also gaze on an object for more information (in the rendered text). {\bf (c)} \& {\bf (d)} The human pinches on the green block and claims it for himself. The robot now projects a faded out green block and re-plans online to use the orange block instead (as evident by pickup arrow that has shifted on the latter at this time). {\bf (e)} Real-time update and rendering of the current state showing status of the plan and objects in the environment. {\bf (f)} The robot completes its new plan using the orange block.
}
\label{replan}
\end{figure*}

\begin{figure*}[!ht]
\centering
\includegraphics[width=0.9\textwidth]{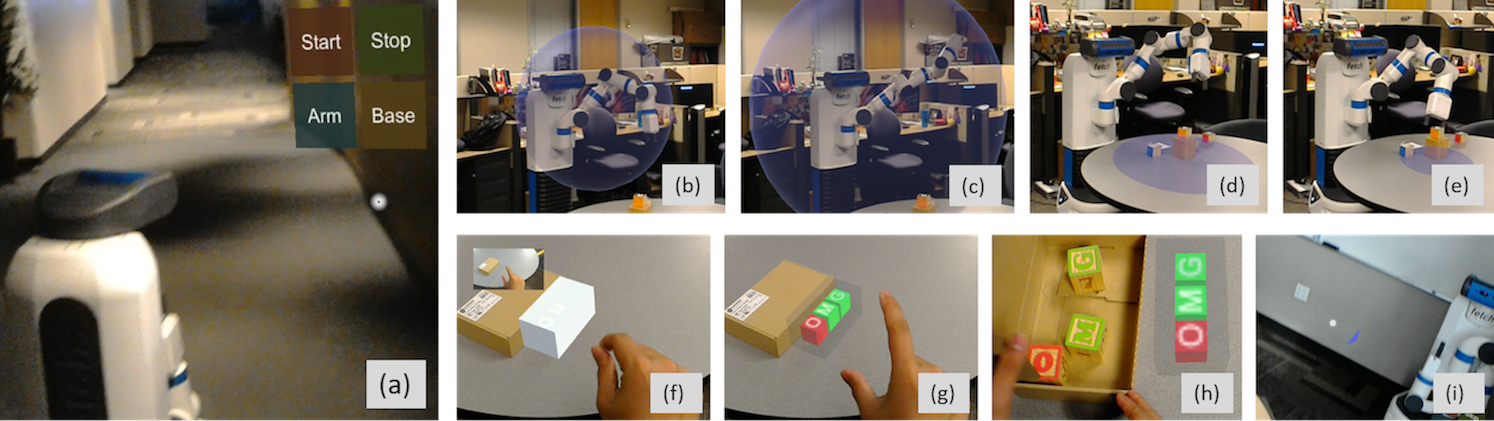}
\caption{
Interactive plan execution using the {\bf (a)} Holographic Control Panel. Safety cues showing dynamic real-time rendering of volume of influence {\bf (b) - (c)} or area of influence {\bf (d) - (e)}, as well as {\bf (i)} indicators for peripheral awareness. Interactive rendering of hidden objects {\bf (f) - (h)} to improve observability and situational awareness in complex workspaces.
}
\label{cues}
\end{figure*}


\subsection*{Modalities of Interaction}

We will now demonstrate different ways augmented reality can improve the human-robot workspace, either by providing a platform for interactive plan execution for online collaboration, or as a means of providing assistive cues to guide the plan execution process. A video demonstrating all these capabilities is available at \url{https://goo.gl/pWWzJb}. 

\subsubsection{Interactive Plan Execution.}

Perhaps the biggest use of AR techniques in the context of planning is for human-in-the-loop plan execution. For example, a robot involved in an assembly task can project the objects it is intending to manipulate into the human's point of view, and annotate them with holograms that correspond to intentions to use or pickup. The human can, in turn, access or claim a particular object in the virtual space and force the robot to re-plan, without there ever being any conflict of intentions in the real space. The humans in the loop can thus not only infer the robot's intent immediately from these holographic projections, but can also interact with them to communicate their own intentions directly and thereby modify the robot's behavior online. The robot can also then ask for help from the human, using these holograms. Figure \ref{replan} shows, in detail, one such use case in our favorite BlocksWorld domain.

The human can go into finer control of the robot by accessing the Holographic Control Panel, as seen in Figure \ref{cues}(a). The panel provides the human controls to start and stop execution of the robot's plan, as well as achieve fine grained motion control of both the base and the arm by making it mimic he user's arm motion gestures on the MoveArm and MoveBase holograms attached to the robot.

\subsubsection{Assistive Cues.}

The use of AR is, of course, not just restricted to procedural execution of plans. It can also be used to annotate the workspace with artifacts derived from the current plan under execution in order to improve the fluency of collaboration. For example, Figure \ref{cues}(b-e) shows the robot projecting its area of influence in its workspace either as a 3D sphere around it, or as a 2D circle on the area it is going to interact with. This is rendered dynamically in real-time based on the distance of the end effector to its center, and to the object to be manipulated. This can be very useful in determining safety zones around a robot in operation. As seen in Figure \ref{cues}(f-i), the robot can also render hidden objects or partially observable state variables relevant to a plan, as well as  indicators to improve peripheral vision of the human, to improve his/her situational awareness.

\section{The Consciousness Cloud}



\begin{figure}[tbp!]
\centering
\includegraphics[width=0.75\columnwidth]{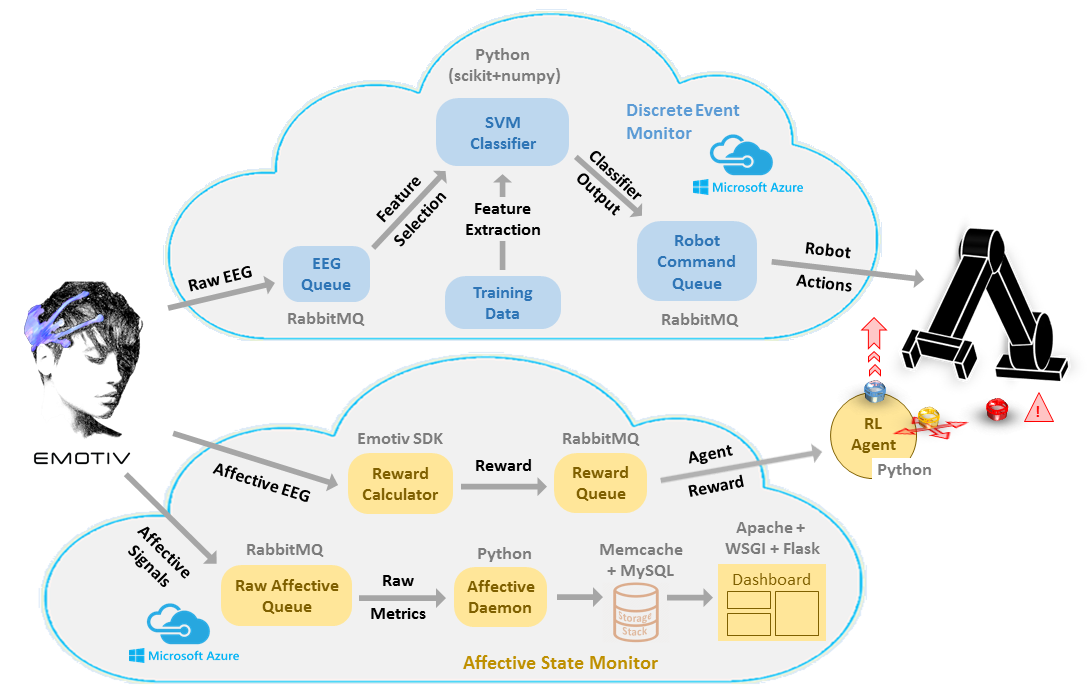}
\caption{Architecture diagram of the Consciousness Cloud.}
\label{cc-arch}
\end{figure}

The Consciousness Cloud has two components - the affective state monitor and the discrete event monitor (as shown in Figure \ref{cc-arch}). In the affective state monitoring system, metrics corresponding to affective signals recorded by the Emotiv EPOC+ headset are directly fed into a rabbitMQ queue, as before, called “Raw Affective Queue” to be used for visualization, and a reward signal (calculated from the metrics) is fed into the “Reward Queue”. The robot directly consumes the “Reward Queue” and the signals that appear during an action execution is considered as the action reward or environment feedback for the AI agent (implementing a reinforcement learning agent). For the discrete event monitoring system, the raw EEG signals from the brain are sampled and written to a rabbitMQ queue called “EEG queue”. This queue is being consumed by our Machine learning or classifier module, which is a python daemon running on a azure server. When this python daemon is spawned it trains an SVM classifier using a set of previously labelled EEG signals. The signals consumed from the queue are first passed through a feature extractor and then the extracted features are used by the SVM to detect specific events (e.g. blinks). For each event a corresponding command is sent to the “Robot Command” queue, which is consumed by the robot. For example, if a STOP command is sent for the blink event, it would cause the robot to halt its current operation.

\begin{figure*}[!ht]
\centering
\includegraphics[width=0.8\textwidth]{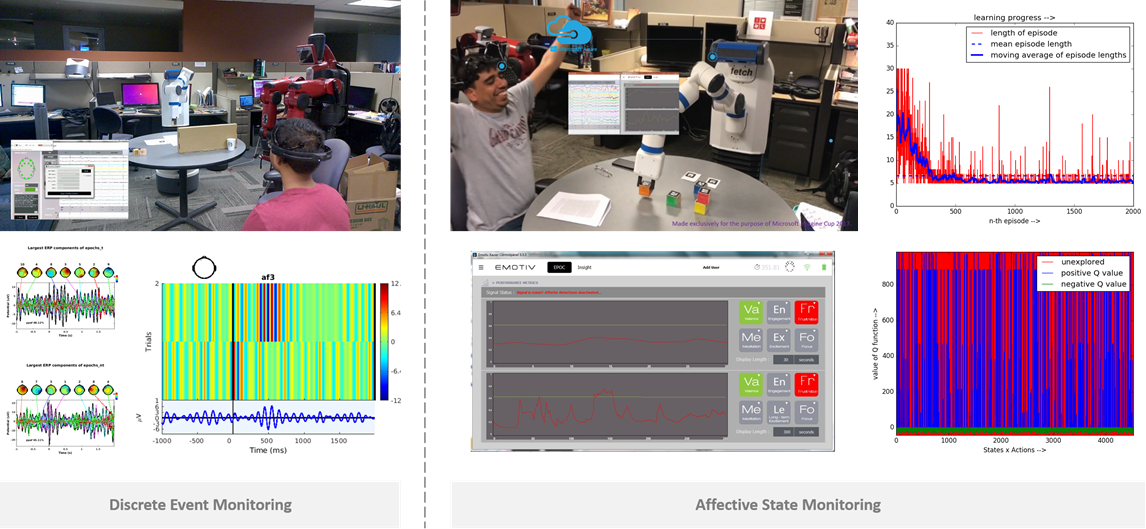}
\caption{
Different modes of EEG feedback - the robot can observe response to discrete events (left - listening for p300s) and listen to longer term affective states of the human (right - a reinforcement learner using stress values as
negative feedback), and use this information to refine its policies.
}
\label{eeg}
\end{figure*}

\subsection*{Modalities of Interaction}

Figure \ref{eeg} demonstrates different ways in which EEG signals can be used to provide closed loop feedback to control the behavior of robots. This can be useful in two ways - either as a means of plan monitoring, i.e. controlling the plan execution process using immediate feedback, or as a reward signal for shaping and refining the policies of a learning agent. A video demonstrating these capabilities is available at  \url{https://goo.gl/6LhKNZ}.

\subsubsection{Discrete Events.}

Discrete events refer to instantaneous or close to instantaneous events, producing certain typical (and easy to classify) signals. 
We identify the following modalities of EEG-based feedback in this regard - (1) Event Related Potentials or ERPs (e.g. p300) that can provide insight into the human's responses like surprise; (2) Affective States like stress, valence, anger, etc. that can provide longer term feedback on how the human evaluates interactions with the robot; and finally (3) Alpha Rhythm that can relate to factors such as task engagement and focus of the human teammate. 
This type of feedback is useful in the online monitoring of the plan execution process by providing immediate feedback on errors or mistakes made by the robot. The video demonstration shows a particular example when the human avoids coming into the harm's way by stops the robot's arm by blinking. Figure \ref{eeg} shows another such use case where the robot is building words (chosen by the human) out of lettered blocks and makes a wrong choice of a letter at some stage - the mistake may be measured as a presence of ERP signal here. The latter has so far gotten mixed results leading us to shift to different EEG helmets (Emotiv Epoc+ lacks electrodes in the central area of the brain where p300s are known to be elicited) for better accuracy.

\subsubsection{Affective States.}

Here, our aim is to train a learning agent to model the preferences of its human teammate by listening to his/her emotions or affective states. We refer to this as {\em affective robotics} (analogous to the field of affective computing). As we mentioned before, the Emotiv SDK currently provides five performance metrics, namely valence/excitement, stress/frustration, engagement, attention, and meditation.
At this time, we have limited ourselves to excitement and stress as our positive ($R^{H+}$) and negative reward signals ($R^{H-}$). We use a linear combination of these two metrics to create a feedback signal that captures the human’s emotional response to a robot’s action.  It is important to note that these signals do not capture the entire reward signal but only capture soft goals or preferences that the robot should satisfy, which means the total reward for the agent is given by $R = R^T + R^H$, where $R^T$ is the reward for the original task.
However, learning this from scratch becomes a hard (as well as somewhat unnecessary if the domain physics is already known) problem given the number of episodes this will require.
Keeping this in mind, we adopt a two staged approach where the learning agent is first trained on the task in isolation without the human in the loop (i.e Q-learning with only $R^T$) so that it can learn a policy that solves the problem ($\pi^T$). Then we use this plan as the initial policy for a new Q-learning agent that considers the full rewards ($R$) with the human in the loop. This {\em ``bootstrapping"} approach should reduce the training time.

The scenario, as seen in Figure \ref{eeg}, involves a workspace that is shared by a robot and a human. The workspace consists of a table with six multicolored blocks. The robot is expected to form a three-block tower from these blocks. As far as the robot is concerned all the blocks are identical and thus the tower can be formed from any of the blocks. The human has a goal of using one of those specific blocks for his/her own purpose. This means whenever the robot uses that specific block it would produce high levels of frustration within the human. The goal of the robot is thus to use this negative reward to update its policy to make sure that it doesn’t use one of the blocks that the human requires.

For the first phase of training, we trained the agent using a simulated model of the task. For the state representation, we used a modified form of the IPC BlocksWorld pddl domain. We used a factored representation of the state with 36 predicates and one additional predicate \texttt{tower3\_formed} to detect task completion. At every step, the agent has access to 50 actions to manipulate the blocks on the table and 80 additional actions \texttt{form3tower} to check for the goal. As for the task rewards, each action is associated with a small negative reward and if the agent achieves the goal it receives a large positive reward. We also introduced an additional reward for every time the number of \texttt{ontable} predicates reduces (which means the agent is forming larger towers) to improve the convergence rate.
We found that the agent converged to the optimal policy (the agent achieves the goal in 5 steps) at around 800 iterations. Figure \ref{eeg} shows the length of the episodes produced after each iteration and the distribution of Q values across the table.
Once the initial bootstrapping process was completed, we used the resultant Q-value table as our input for the second phase of the learning, as seen in the video demonstration. While there are some issues with convergence that are yet to be resolved, initial results showing the robot exploring new policies using the stress signals are quite exciting.

\section{Conclusions \& Future Work}

In conclusion, we presented two approaches to improve collaboration among humans and robots from the perspective of task planning, either in terms of an interactive plan execution process or in gathering feedback to inform the human-aware decision making process. To this end, we discussed the use of holograms as a shared vocabulary for effective communication in an augmented workspace. We also discussed the use of EEG signals for immediate monitoring, as well as long term feedback on the human response to the robot, which can be used by a learning agent to shape its policies towards increased human-awareness. 
Such modes of interaction opens up several exciting avenues of research. We mention a few of these below.

\subsubsection{Closing the planning-execution loop}

The ability to project intentions and interact via those projections may be considered in the plan generation process itself - e.g. the robot can prefer a plan that is easier to project to the human for the sake of smoother collaboration. This notion of projection-aware task or motion planning adds a new dimension to the area of human-aware planning.

A holographic vocabulary also calls for the development of representations - PDDL3.x - that can capture complex interaction constraints modeling not just the planning ability of the agent but also its interactions with the human. Further, such representations can be \emph{learned} to generalize to methods that can, given a finite set of symbols or vocabulary, compute domain independent projection policies that decide what and when to project to reduce cognitive overload on the human.

\subsubsection{ERP and timed events} Perhaps the biggest challenge towards adopting ERP feedback over a wide variety of tasks is the reliance of detecting these signals on the exact time of occurrence of the event. Recent advancements in machine learning techniques can potentially allow windowed approaches to detect such signals from raw data streams.

\subsubsection{Evaluations} 
While preliminary studies with fellow graduate student subjects have been promising, we are currently working towards systematic evaluation of our system under controlled conditions, complying with the ISO 9241-11:1998 standards, targeted at professionals who are engaged in similar activities repeatedly over prolonged periods. This is essential in evaluating such systems since the value of information in projections is likely to reduce significantly with expertise and experience. 









\bibliographystyle{IEEEtran}
\bibliography{IEEEabrv,bib}

\end{document}